\documentclass[10pt,twocolumn,letterpaper]{article}

\usepackage{cvpr}
\usepackage{times}
\usepackage{epsfig}
\usepackage{graphicx}
\usepackage{amsmath}
\usepackage{amssymb}

\usepackage{bm}
\usepackage{booktabs}
\usepackage{courier}

\usepackage[pagebackref=true,breaklinks=true,letterpaper=true,colorlinks,bookmarks=false]{hyperref}

\cvprfinalcopy 

\def\cvprPaperID{****} 

\begin{document}

\title{Knowledge Aided Consistency for Weakly Supervised Phrase Grounding}

\author{
   Kan Chen\qquad Jiyang Gao\qquad Ram Nevatia \\
   Institute for Robotics and Intelligent Systems, University of Southern California \\
   Los Angeles, CA 90089, USA \\
   {\tt\small\{kanchen|jiyangga|nevatia\}@usc.edu}
}

\maketitle

\begin{abstract}
Given a natural language query, a phrase grounding system aims to localize mentioned objects in an image. 
In weakly supervised scenario, mapping between image regions (i.e.,~proposals) and language is not available in the training set.
Previous methods address this deficiency by training a grounding system via learning to reconstruct language information contained in input queries from predicted proposals.
However, the optimization is solely guided by the reconstruction loss from the language modality, and ignores rich visual information contained in proposals and useful cues from external knowledge.
In this paper, we explore the consistency contained in both visual and language modalities, and leverage complementary external knowledge to facilitate weakly supervised grounding.
We propose a novel Knowledge Aided Consistency Network (KAC Net) which is optimized by reconstructing input query and proposal's information. 
To leverage complementary knowledge contained in the visual features, we introduce a Knowledge Based Pooling (KBP) gate to focus on query-related proposals.
Experiments show that KAC Net provides a significant improvement on two popular datasets.
\end{abstract}

\section{Introduction}\label{sec: intro}

Given an image and a natural language query, phrase grounding aims to localize objects mentioned by the query. 
It is a fundamental building block for many high-level computer vision tasks such as image retrieval~\cite{Chen_2017_CVPR}, image QA~\cite{chen2015abc,gan2017stylenet,gan2017vqs} and video QA~\cite{gao2018motion,gao2017tall}. 
Traditionally, training a good phrase grounding system requires large amounts of manual annotations indicating the mapping between input queries and mentioned objects in images; these are time-consuming to acquire and suffer from potential human errors.
This motivates us to address the problem of training a grounding system by weakly supervised training data where objects of interest are mentioned in language queries but are not delineated in images.

\begin{figure}[t]
\includegraphics[width=3.3in]{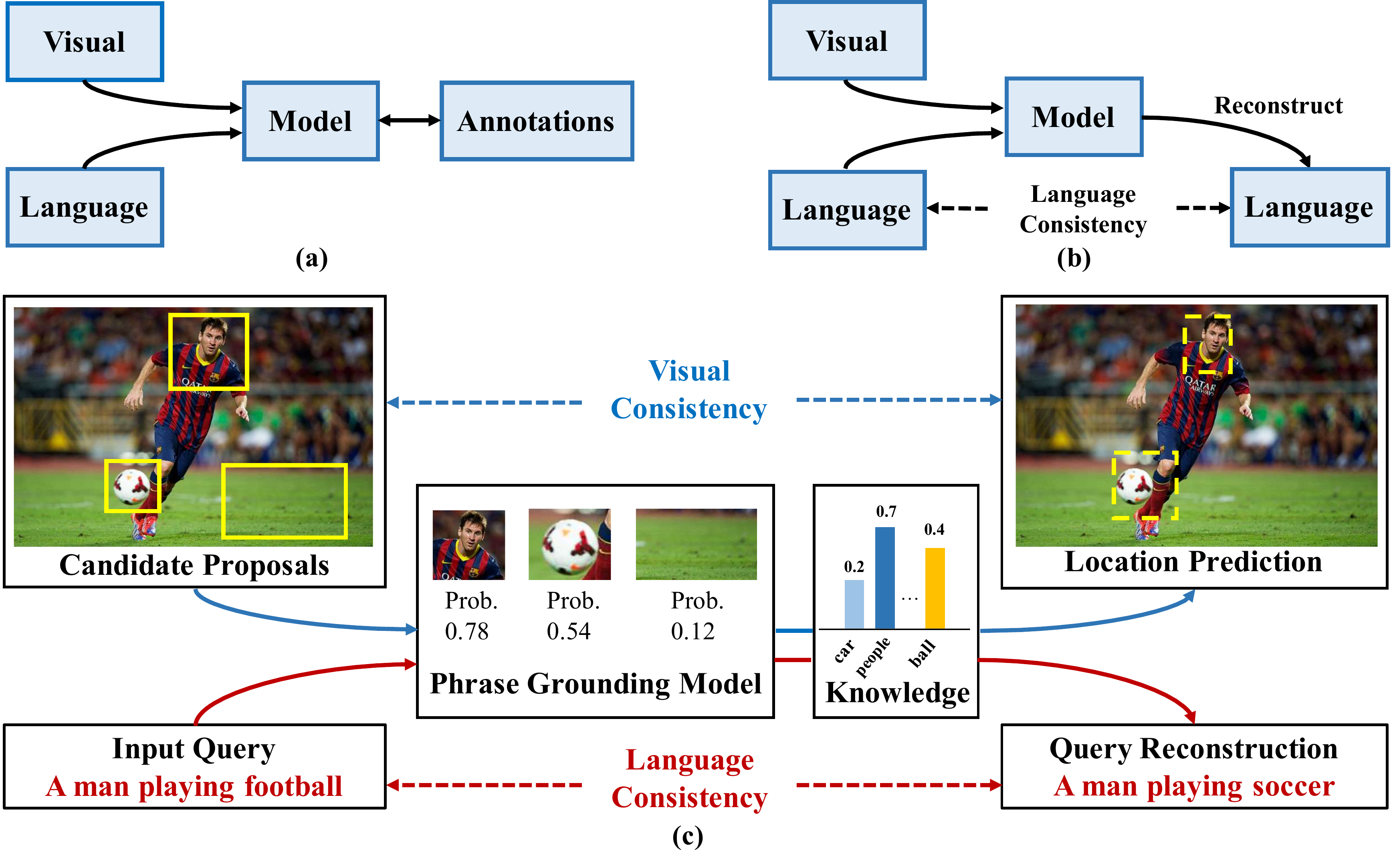}
\centering
\caption{(a) supervised grounding systems, (b) state-of-the-art weakly supervised grounding systems guided by language consistency, (c) KAC Net applies both visual and language consistency and leverages complementary knowledge from the visual feature extractor to facilitate weakly supervised grounding.}\label{fig: intro}\vspace{-3mm}
\end{figure}

Phrase grounding is difficult as both visual and language modalities are ambiguous and we need to reason about both to find their correspondences.
To address this problem, typically a proposal generation system is applied to the input image to produce a set of candidate regions (\emph{i.e.}, proposals). 
Phrase grounding task is then treated as a retrieval problem to search the most query-related proposals.
Based on this, attention mechanisms~\cite{chen2017msrc,kan2017qrc,rohrbach2016grounding,xiao2017weakly} are learned to adaptively attend to mentioned objects for input queries.

Training a phrase grounding system with weakly supervised data brings additional challenge as no direct mappings between the two modalities are provided. 
Consider Fig.~\ref{fig: intro}(c) where we encode the query as an embedding vector and extract visual features for a set of object proposals from the image. 
To find correct mappings between the query and the proposals,~\cite{rohrbach2016grounding} proposes to associate the query with successive proposals; once a proposal is selected, a phrase is reconstructed from it and evaluated for language consistency with the input query. 
~\cite{xiao2017weakly} adopts continuous attention maps and explores to reconstruct the structure of input query as well as its context.

We introduce two new concepts to overcome challenges of weakly supervised training. First is that pre-trained, fixed category detectors can provide useful knowledge in selecting the proposals that should be attended to. Second is that the detector knowledge enables us to evaluate visual consistency, in addition to language consistency. This knowledge also helps improve language consistency analysis.

We observe that if a pre-trained Convolutional Neural Network (CNN) (\emph{e.g.}, VGG~\cite{Simonyan14c}) is applied to extract visual features for proposals, it can also naturally produce a probability distribution of the categories of the proposals, as this is the task that the network was trained on (\emph{e.g.} MSCOCO~\cite{lin2014microsoft} classification). 
This free distribution can be treated as complementary external knowledge to filter out, or downweight,  proposals that are unrelated to the query.
For example, in Fig.~\ref{fig: intro}(c), given a query ``a man playing football'', a pre-trained VGG network can provide useful hints for candidate proposals by predicting whether a proposal corresponds to a high probability ``people'' detection. 

Use of external knowledge in language consistency is straight-forward; features for reconstruction can be modified by the detection probabilities. Task of evaluating visual consistency is more difficult; a direct analogy to language consistency would be to convert visual proposal to words and reconstruct image patches. 
Instead, we propose to predict object locations from query and visual features to match the goal of phrase grounding. 
This process would be not possible without the aid of external knowledge that helps focus on the possible related proposals for prediction.

In implementation, we construct a novel Knowledge Aided Consistency Network (KAC Net) which consists of two branches: a visual consistency branch and a language consistency branch.
These two branches are joined by a shared multimodal subspace where the attention model is applied.
To leverage complementary knowledge from visual feature extractor, we propose a novel Knowledge Based Pooling (KBP) gate to focus on query-related proposals for visual and language reconstruction.

We evaluate KAC Net on two grounding datasets: Flickr30K Entities~\cite{plummer2015flickr30k} and Referit Game~\cite{KazemzadehOrdonezMattenBergEMNLP14}. 
Flickr30K Entities contains more than 30K images and 170K query phrases, while Referit Game has 19K images referred by 130K query phrases. 
We ignore bounding box annotations during training in weakly supervised scenario.
Experiments show KAC Net outperforms state-of-the-art methods by a large margin on both two datasets, with more than 9\% increase on Flickr30K Entities and 5\% increase on Referit Game in accuracy.

Our contributions are twofold: First, we leverage complementary knowledge to filter out unrelated proposals and provide direct guidance. Second, we propose a visual consistency to boost grounding performance. In the following paper, we first discuss related work in Sec.~\ref{sec: related work}. More details of KAC Net are provided in Sec.~\ref{sec: method}. Finally we analyze and compare KAC Net with other approaches in Sec.~\ref{sec: exps}.

\section{Related Work}\label{sec: related work}
\textbf{Phrase grounding} requires learning similarity between visual and language modalities.
Karpathy \emph{et al.}~\cite{karpathy2014deep} first align sentence fragments and image regions in a subspace, and later apply a bi-directional RNN for multimodal alignment in~\cite{karpathy2015deep}.  
Hu \emph{et al.}~\cite{hu2016natural} employ a 2-layer LSTM to rank proposals based on encoded query and visual features. 
Rohrbach \emph{et al.}~\cite{rohrbach2016grounding} employ a latent attention network conditioned on query which ranks proposals in weakly supervised scenario. 
Recently, Plummer \emph{et al.}~\cite{plummer2015flickr30k} augment the CCA model~\cite{plummer2014flickr30k} to leverage extensive linguistic cues in the phrases. 
Chen \emph{et al.}~\cite{chen2017msrc} introduce regression mechanism in phrase grounding to improve proposals' quality.
Xiao \emph{et al.}~\cite{xiao2017weakly} leverage query's language structural information to guide the learning of phrase grounding model in weakly supervised scenario.
Chen \emph{et al.}~\cite{kan2017qrc} apply reinforcement learning techniques to leverage context information. 
In this paper, we explore consistency in visual and language modalities and leverage complementary knowledge to further boost performance of weakly supervised grounding.

\begin{figure*}[t]
\includegraphics[width=6.8in]{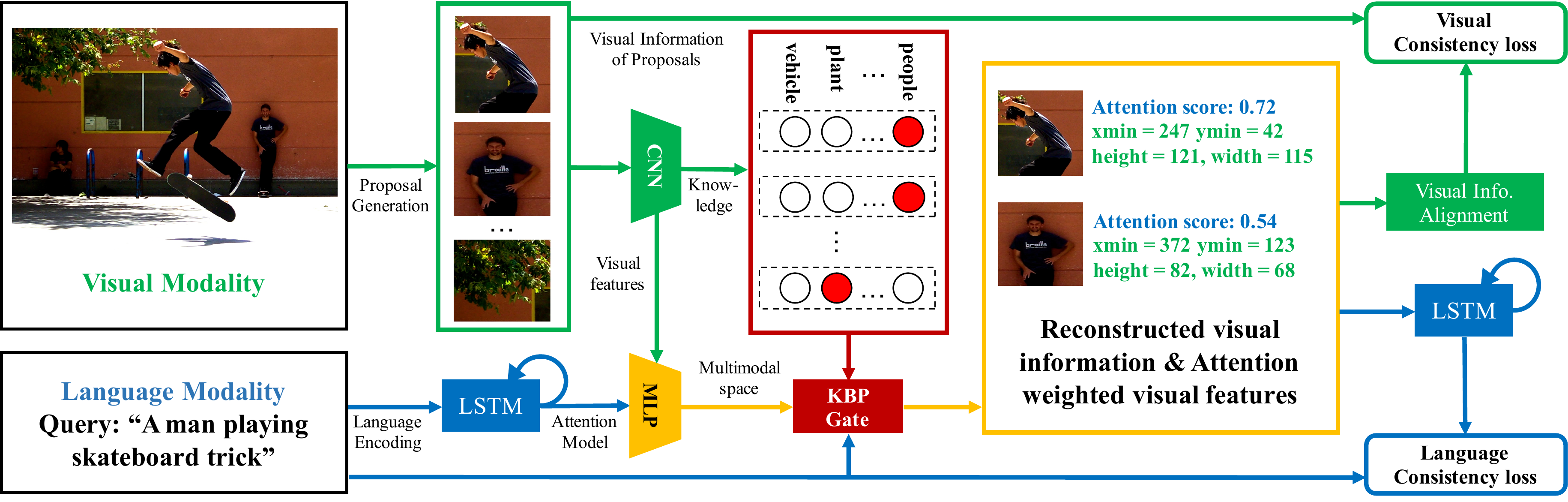}
\centering
\caption{Knowledge Aided Consistency Network (KAC Net) consists of a visual consistency branch and a language consistency branch. Visual consistency branch aims at predicting and aligning query-related proposals' location parameters conditioned on the input query. Language consistency branch attempts to reconstruct input query from query-related proposals. To provide guidance in training and testing, a Knowledge Based Pooling (KBP) gate is applied to filter out unrelated proposals for both branches.}\label{fig: framework}
\end{figure*}

\textbf{Weakly supervised learning} is a method aims at learning a model without heavy manual labeling work. 
It is widely used in different computer vision tasks. 
Crandall \emph{et al.}~\cite{crandall2006weakly} leverage the class labeling to learn a part-based spatial model without detailed annotation of object location and spatial relationship.
Maxime \emph{et al.}~\cite{oquab2015object} propose to learn the interaction between human and objects purely from action labeling for still images. 
Recently, Prest \emph{et al.}~\cite{prest2012weakly} apply a deep convolutional neural network and its score maps to address object localization with image level class labels.
For phrase grounding task, Rohrbach \emph{et al.}~\cite{rohrbach2016grounding} propose to adopt an attention model which is optimized by learning to reconstruct query's information, and avoids human labeling for object locations for each query in the training set.
Based on this, Xiao \emph{et al.}~\cite{xiao2017weakly} leverage a continuous attention map and explore detailed structural reconstruction of language modality.
Inspired by the success of weakly supervised learning, we propose to apply another visual consistency to further boost performance.


\textbf{Knowledge transfer} is a technique widely used for tasks in different domains. 
Hinton \emph{et al.}~\cite{hinton2015distilling} propose to compress knowledge learned from one model into another one which is too computationally expensive to train.
Inspired by this, Aytar \emph{et al.}~\cite{aytar2016soundnet} apply visual knowledge to train a sound classification network.
Owens \emph{et al.}~\cite{owens2016ambient} use ambient sound information to train an object detection network.
Lin \emph{et al}.~\cite{lin2016leveraging} leverage knowledge learned in Visual Question Answering (VQA) task in image retrieval.
Zhang \emph{et al.}~\cite{zhang2017visual} apply knowledge learned in image captioning and VQA to train a network detecting visual relation in images.
For phrase grounding, we propose to leverage knowledge learned from pre-trained deep neural network to filter out unrelated proposals for visual consistency.

\section{KAC Network}\label{sec: method}
KAC Net consists of two branches: a visual consistency branch and a language consistency branch which reconstructs visual and language information respectively.
The two branches are joined in a shared multimodal subspace, where an attention model is applied to attend on mentioned objects based on query's semantics.
To leverage external knowledge from pre-trained CNN feature extractor, a Knowledge Based Pooling (KBP) gate is proposed to select query-related proposals.
KAC Net is trained end-to-end, with both visual and language consistency restriction to guide the training.

We first introduce the framework of KAC Net, followed by the details of KBP gate.
Then we illustrate how KBP is applied to facilitate the optimization of visual and language consistency branches.
Finally, more details of training and inference are provided.

\subsection{Framework}
The goal of KAC Net is to localize the mentioned object $y$ given a query phrase $q$ and an image $x$.
To address the problem, a set of $N$ proposals $\{r_i\}$ are generated via an object proposal generation system. 
An attention model is then applied to attend on the proposal $r^q$ which contains the mentioned object $y$ based on the semantics of query $q$.



In weakly supervised scenario, the mapping between query $q$ and the location of mentioned object $y$ is not provided. 
To learn the attention model, we adopt visual and language consistency and construct two branches respectively.
For language consistency, a reconstruction model is applied to reconstruct input query $q$ given the query-related proposals predicted by the attention model.
According to the language consistency, the reconstructed query should be consistent with the input. 
A language consistency loss $\mathcal{L}_{lc}$ is generated by comparing the reconstructed and original queries. 

For visual consistency, we propose to reconstruct visual information for query-related proposals.
Since the goal of phrase grounding is to predict mentioned object's location, we choose to predict candidate proposals' location parameters conditioned on the input query. 
Similar to language consistency, visual consistency requires that the predicted parameters should recover each proposal's location. 
Based on this, a visual consistency loss $\mathcal{L}_{vc}$ is produced by calculating the difference between the predicted and original proposals' location parameters.
 
To leverage rich image features and available fixed category classifiers, we apply KBP to encode knowledge provided by CNN and weight each proposal's importance in visual and language consistency.
The objective of KAC Net can be written as
\begin{equation}\label{equ: KAS_obj}
\arg\min_\theta\sum_q(\mathcal{L}_{lc}^k + \lambda\mathcal{L}_{vc}^k)+\mu\mathcal{L}_{reg}
\end{equation}
where $\theta$ denotes the parameters to be optimized. $\mathcal{L}_{lc}^k$ is the reconstruction loss from language consistency branch and $\mathcal{L}_{vc}^k$ is the reconstruction loss from visual consistency branch (superscript ``$k$'' refers to KBP). 
$\mathcal{L}_{reg}$ is a weight regularization term. $\lambda$, $\mu$ are hyperparameters. 

\subsection{Knowledge Based Pooling (KBP)}

We apply a pre-trained CNN to extract visual feature $\mathbf{v}_i$ for a proposal $r_i$, and predict a probability distribution $\bm{p}_i$ for its own task, which provides useful cues to filter out unrelated proposals. 

\begin{figure}[t]
\includegraphics[width=3.2in]{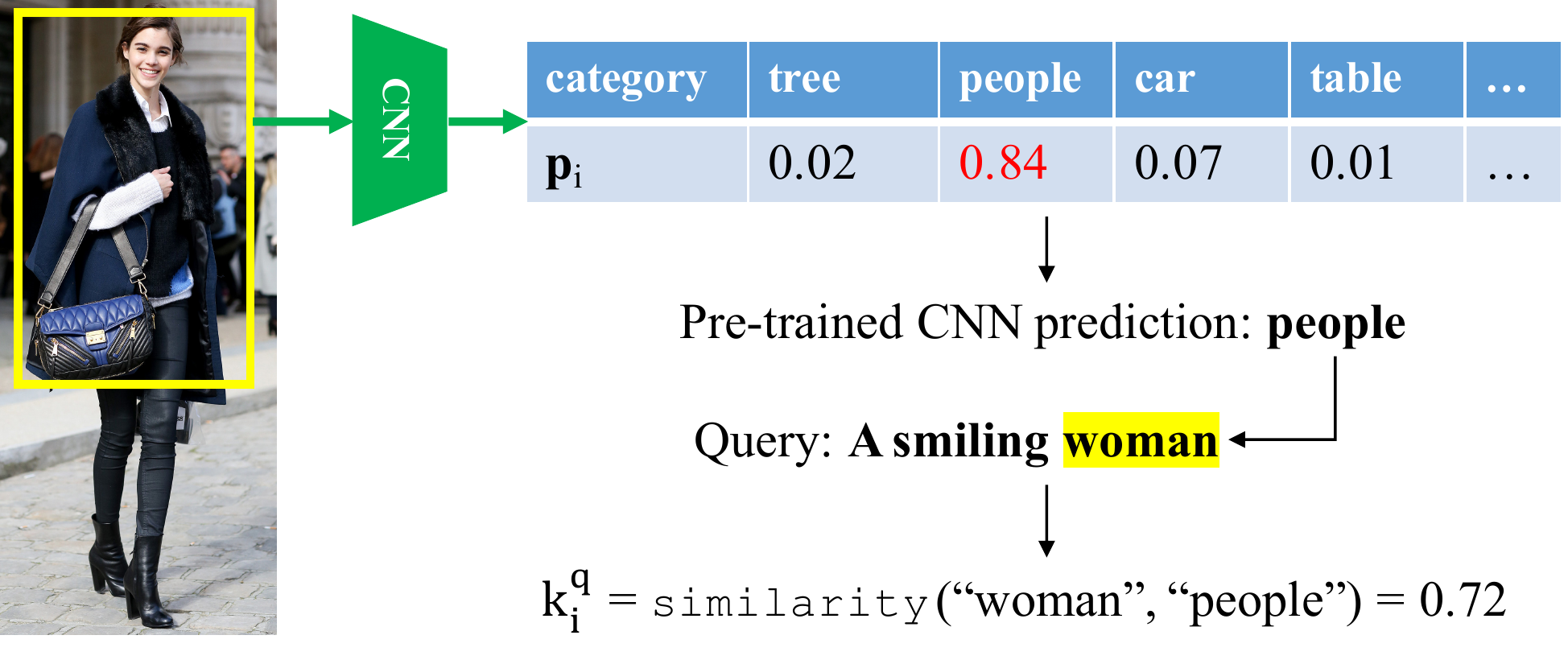}
\centering
\caption{A pre-trained CNN always predicts a probability distribution for its own task. We leverage the most probable category predicted by CNN and calculate the word similarity between noun words in the query as knowledge $k_i^q$}\label{fig: kbp}
\end{figure}

To encode this knowledge, we first parse the language query and retrieve all the noun words via a Natural Language Processing (NLP) parser. 
For each proposal's distribution $\bm{p}_i$, we select the most probable class with the highest probability.
The knowledge $k_i^q$ for proposal $r_i$ is then calculated as the word similarity between the name of this class and noun words in the query (Fig.~\ref{fig: kbp}).
If a query contains multiple noun words, we average all the calculated similarities as the knowledge $k_i^q$, which can be written as

\begin{equation}\label{equ: knowledge calc}
k_i^q = \frac{1}{N_q}\sum_{j=1}^{N_q}\texttt{sim}(C_i^*, w_j^q)
\end{equation}
where $C_i^*$ is the predicted class name for proposal $r_i$, $w_j^q$ is the $j$-th word of all the $N_q$ noun words in the query $q$. \texttt{sim} is a  function measuring the similarity between two words.

In the training stage, knowledge $k_i^q$ functions as a ``pooling'' gate which helps visual (Sec.~\ref{ssec: vmv}) and language (Sec.~\ref{ssec: lml}) consistency branches select and reconstruct reliable candidate proposals.
In the test stage, knowledge $k_i^q$ filters out unrelated proposals and increases the chance of finding the proposal containing the mentioned object (Sec.~\ref{ssec: test}). 

\subsection{Visual Consistency}\label{ssec: vmv}
The goal of visual consistency is to optimize the attention model via learning to predict location information contained in query-related proposals. 
Through predicting location information conditioned on the input query, we expect to learn a better correlation between language and visual modalities.
In weakly supervised scenario, no annotations are available to indicate the identity of query-related proposal. 
Instead, we use KBP's knowledge $k_i^q$ to provide guidance during training.
We expect that knowledge $k_i^q$ provides a higher score when a proposal $r_i$ is query related.
Thus, KBP can be applied to adaptively weight each proposal's visual consistency loss conditioned on query $q$.

 
In implementation, we first apply a Long Short-Term Memory (LSTM)~\cite{hochreiter1997long} model to encode input query $q$ into an embedding vector $\mathbf{q}\in\mathbb{R}^{d_q}$.
A pre-trained CNN is employed to extract visual feature $\mathbf{v}_i\in\mathbb{R}^{d_v}$ for each proposal $r_i$, and global visual feature $\mathbf{v}\in\mathbb{R}^{d_v}$ for input image $x$.
The attention model then concatenates the embedding vector $\mathbf{q}$, image global feature $\mathbf{v}$ with each of the proposal's feature $\mathbf{v}_i$ and projects them into an $m$-dimensional subspace. 
A multimodal feature $\mathbf{v}_i^{q}$ is calculated as
\begin{equation}\label{equ: multimodal space}
\mathbf{v}_i^{q} = \varphi(\mathbf{W}_m(\mathbf{q}||\mathbf{v}||\mathbf{v}_i) + \mathbf{b}_m)
\end{equation}
where $\mathbf{W}_m\in\mathbb{R}^{m\times(d_q+2d_v)}$, $\mathbf{b}_m\in\mathbb{R}^m$ are projection parameters. $\varphi(.)$ is a non-linear activation function. ``$||$'' denotes a concatenation operator.

After projecting into the multimodal subspace, the attention model predicts a 5D vector $\mathbf{s}^p\in\mathbb{R}^5$ via a fully connected (fc) layer (superscript ``p'' denotes prediction).
\begin{equation}\label{equ: att score}
\mathbf{s}_i^p = \mathbf{W}_s\mathbf{v}_i^q + \mathbf{b}_s
\end{equation}
where $\mathbf{W}_s\in\mathbb{R}^{5\times m}$ and $\mathbf{b}_s\in\mathbb{R}^5$ are projection parameters. 
The first element in $\mathbf{s}_i^p$ estimates the confidence of $r_i$ being relevant to input query $q$, and the next four elements represent the predicted location parameters for each proposal.

We compare the predicted location parameters with original proposal's parameters $\mathbf{t}_i\in\mathbb{R}^4$ and calculate the regression loss
\begin{equation}\label{equ: regression loss}
d_i = \frac{1}{4}\sum_{j=0}^3f(|\mathbf{t}_i[j]-\mathbf{s}_i^p[j+1]|)
\end{equation}
where $f(.)$ is the smooth L1 loss function: $f(x) = 0.5x^2$ $(|x|<1)$, and $f(x) = |x|-0.5 (|x|\geq1)$.
The location parameters $\mathbf{t}_i$ are in the form $[x_{i1}/w, y_{i1}/h, x_{i2}/w, y_{i2}/h]-0.5$, where $x_{i1}, x_{i2}$ is the minimum and maximum $x$-axis location of proposal $r_i$, and $y_{i1}, y_{i2}$ is the minimum and maximum $y$-axis location.

Aided by KBP gate, we weight each proposal's regression loss $d_i$ based on the predicted confidence $\mathbf{s}_i^p[0]$ and knowledge $k_i^q$. The visual consistency loss $\mathcal{L}_{vc}^k$ is calculated as 
\begin{equation}\label{equ: vmv loss}
\mathcal{L}_{vc}^k = \sum_{i=1}^{N}\sigma(k_i^q)\phi(\mathbf{s}_i^p[0])d_i
\end{equation}
where $\phi(.), \sigma(.)$ denotes a softmax function and a sigmoid function respectively.

\subsection{Language Consistency}\label{ssec: lml}
The goal of language consistency is to optimize the attention model via learning to reconstruct input query $q$ with a language consistency constraint.

In implementation, after the attention model predicting each proposal's confidence of being relevant to query $q$ ($\mathbf{s}_i^p[0]$ in Eq.~\ref{equ: att score}), we adopt a similar structure in~\cite{rohrbach2016grounding} to weight each proposal's visual feature $\mathbf{v}_i$ and project them into a reconstruction subspace.
Different from~\cite{rohrbach2016grounding}, we introduce KBP gate into the language consistency branch to further down-weight unrelated visual features' contribution.
Thus, the knowledge conditioned reconstruction feature is calculated as
\begin{equation}\label{equ: vis recons feat}
\mathbf{v}_{att}^k = \mathbf{W}_a\left(\sum_{i=1}^N\sigma(k_i^q)\phi(\mathbf{s}_i^p[0])\mathbf{v}_i\right)+\mathbf{b}_a
\end{equation}
where $\mathbf{W}_a\in\mathbb{R}^{d_r\times d_v}$, $\mathbf{b}_a\in\mathbb{R}^{d_r}$ are projections parameters to be optimized. Other notations are the same as Eq.~\ref{equ: vmv loss}. 

The reconstruction visual feature $\mathbf{v}_{att}^k$ is then treated as the initial state of a decoding LSTM, which predicts a sequence of probability $\{\bm{p}_{\hat{q}}^t\}$ indicating the selection of words in each time step $t$ of reconstructed query $\hat{q}$.
With the ground truth of input query $q$ (selection of words $w_t$ in each time step $t$), the language reconstruction loss $\mathcal{L}_{lc}^k$ is the average of cross entropy for the sequence $\{\bm{p}_{\hat{q}}^t\}$.
\begin{equation}\label{equ: lml loss}
\mathcal{L}_{lc}^k = -\frac{1}{T}\sum_{t=1}^T\log(\bm{p}_{\hat{q}}^t[w_t])
\end{equation}
where $T$ is the length of input query $q$. 

\subsection{Training \& Inference}\label{ssec: test}
In training stage, the parameters to be optimized include parameters in encoding and decoding LSTM and the projection parameters in Eq.~\ref{equ: multimodal space},~\ref{equ: att score},~\ref{equ: vis recons feat}. 
We regularize the weights of projection parameters, which is the sum of $\ell_2$ norm of these parameters ($\mathcal{L}_{reg}$). 
Same as~\cite{rohrbach2016grounding}, we select 100 proposals produced by proposal generation systems ($N=100$).
The rectified linear unit (ReLU) is selected as the non-linear activation function $\varphi$.
KAC Net is trained end-to-end using the Adam~\cite{kingma2014adam} algorithm.

In test stage, we feed the query $q$ into the trained KAC Net, and select the most related proposal based on the confidence $\{\mathbf{s}_i^p[0]\}$ generated by the attention model (Eq.~\ref{equ: att score}) and external knowledge $k_i^q$. 
The final prediction is given as (notations are the same in Eq.~\ref{equ: vmv loss}):
\begin{equation}\label{equ: inference}
r_{j^*} \text{,\ \ s.t.\ \ } j^* = \arg\max_i\mathbf\{\phi(\mathbf{s}_i^p[0])\sigma(k_i^q)\}
\end{equation}

\section{Experiment}\label{sec: exps}

We evaluate KAC Net on Flickr30K Entities~\cite{plummer2015flickr30k} and Referit Game~\cite{KazemzadehOrdonezMattenBergEMNLP14} datasets in weakly supervised grounding scenario.

\subsection{Datasets}
\textbf{Flickr30K Entities}~\cite{plummer2015flickr30k}: 
There are 29783, 1000, 1000 images in this dataset for training, validation and testing respectively.
Each image is associated with 5 captions, with 3.52 query phrases in each caption on average (360K query phrases in total).
The vocabulary size for all these queries is 17150.
We ignore the bounding box annotations of these two datasets in weakly supervised scenario.

\textbf{Referit Game}~\cite{KazemzadehOrdonezMattenBergEMNLP14}: 
There are 19,894 images of natural scenes in this dataset, with 96,654 distinct objects in these images. 
Each object is referred to by 1-3 query phrases (130,525 in total).
There are 8800 unique words among all the phrases, with a maximum length of 19 words.

\subsection{Experiment Setup}\label{sec: exp setup}

\textbf{Proposal generation.} We adopt Selective Search~\cite{uijlings2013selective} for Flickr30K Entities~\cite{plummer2015flickr30k} and EdgeBoxes~\cite{zitnick2014edge} for Referit Game~\cite{KazemzadehOrdonezMattenBergEMNLP14} to generate proposals as grounding candidates for fair comparison with~\cite{rohrbach2016grounding} on these two datasets.

\textbf{Visual feature representation.} Same as~\cite{rohrbach2016grounding}, we choose a VGG Network~\cite{Simonyan14c} finetuned by Fast-RCNN~\cite{girshickICCV15fastrcnn} on PASCAL VOC 2007~\cite{pascal-voc-2007} to extract visual features for Flickr30K Entities, which are denoted as ``VGG\textsubscript{det}''.
Besides, we follow~\cite{rohrbach2016grounding} and apply a VGG Network pre-trained on ImageNet~\cite{deng2009imagenet} to extract visual features for both Flickr30K Entities and Referit Game datasets, which are denoted as ``VGG\textsubscript{cls}''.
Both ``VGG\textsubscript{cls}'' and ``VGG\textsubscript{det}'' features are 4096D vectors ($d_v=4096$).

\textbf{Knowledge representation.} To parse different queries, we use the Stanford NLP parser~\cite{manning2014stanford} to extract noun words in each query.
We then extract probability distributions of ``VGG\textsubscript{det}'' features in  MSCOCO~\cite{lin2014microsoft} image classification task for all proposals (\#classes=90).
The similarity between noun words in queries and class names are calculated as the cosine distance via a word2vec program~\cite{mikolov2013efficient}.
We extract probability distributions in PASCAL VOC 2007 classification task~\cite{pascal-voc-2007} (\#classes=20). 
Results of different knowledge facilitation is provided in Sec.~\ref{sec: flickr exp} and~\ref{sec: referit exp}.

\textbf{KBP gate.} For KBP gate, we adopt a soft version and a hard version. 
Soft KBP applies the sigmoid function to transform external knowledge $k_i^q$ into probability to directly weight each proposal, while hard KBP applies thresholding to force probability being either 0 or 1 for each proposal (\emph{i.e.}, $k_{ih}^{q}=\delta(k_{is}^q\geq t)$, $\delta$ is an indicator function, subscripts ``h'', ``s'' denote hard KBP and soft KBP respectively). 

In experiments, we set the threshold $t$ as 0.3 for Flickr30K Entities and 0.1 for Referit Game.
For hard KBP, if a query's knowledge scores are 0 for all proposals (\emph{i.e.} $k_{ih}^q=0, \forall i$), we set them to be all 1 for language reconstruction in Eq.~\ref{equ: vis recons feat}; otherwise, reconstruction features $\mathbf{v}_{att}^k$ provides no information to reconstruct the input query.

\textbf{Model initialization.} Following same settings as in~\cite{rohrbach2016grounding}, input queries are encoded through an LSTM model, and the query embedding vector $\mathbf{q}$ is the last hidden state from LSTM ($d_q = 512$).
All fc layers are initialized by Xavier method~\cite{glorot2010understanding} and all convolutional layers are initialized by MSRA method~\cite{he2015delving}. 
We introduce batch normalization layers after projecting visual and language features in Eq.~\ref{equ: multimodal space}.

During training, we set the batch size as 40. The dimension of multimodal features $\mathbf{v}_i^q$ is set to $m=128$ (Eq.~\ref{equ: multimodal space}). Hyperparameter $\mu$ for weight regularization is 0.005 and $\lambda$ for visual reconstruction loss is 10.0 in Eq.~\ref{equ: KAS_obj}. Analysis of hyperparameters is provided in the supplemental file. 

\textbf{Metric.} Same as~\cite{rohrbach2016grounding}, we adopt accuracy as the evaluation metric, which is defined as the ratio of phrases for which the regressed box overlaps with the mentioned object by more than 50\% Intersection over Union (IoU).

\textbf{Compared approach.} We choose GroundeR~\cite{rohrbach2016grounding} as the compared approach, which achieves state-of-the-art performance on both Flickr30K Entities and Referit Game datasets.

\begin{table}[t]
  \centering
  \begin{tabular}{lc} \toprule
  Approach & Accuracy (\%) \\ \midrule
  \textbf{Compared approaches} & \\
  GroundeR (LC) (VGG\textsubscript{cls})~\cite{rohrbach2016grounding} & 24.66 \\
  GroundeR (LC) (VGG\textsubscript{det})~\cite{rohrbach2016grounding} & 28.93 \\ \midrule
  \textbf{Our approaches} & \\
  VC + Hard KBP (VGG\textsubscript{det})  & 28.58 \\
  VC + Soft KBP (VGG\textsubscript{det}) & 30.60 \\     
  LC + Hard KBP (VGG\textsubscript{det}) & 32.17 \\ 
  LC + Soft KBP (VGG\textsubscript{det}) & 34.31 \\ 
  KAC Net + Hard KBP (VGG\textsubscript{det}) & 37.41 \\
  KAC Net + Soft KBP (VGG\textsubscript{det}) & \textbf{38.71} \\
  \bottomrule
  \end{tabular}
  \vspace{1.5mm}
\caption{Different models' performance on Flickr30K Entities. We explicitly evaluate performance of visual consistency (VC), language consistency (LC) branches with Hard and Soft KBP Gates. We leverage knowledge from MSCOCO~\cite{lin2014microsoft} classification task.}\label{tab: flickr30k res}
\end{table}

\begin{table}[t]
  \centering
  \begin{tabular}{|l|c|c|} \hline
  Knowledge & PASCAL VOC~\cite{pascal-voc-2007} & MSCOCO~\cite{lin2014microsoft} \\ \hline
  Hard KBP & 35.24 & 37.41 \\ \hline
  Soft KBP & 36.14 & 38.71
  \\\hline 
  \end{tabular}
  \vspace{1.5mm}
\caption{Comparison of KAC Net using different KBP gates and external knowledge on Flickr30k Entities. Accuracy is in \%.}\label{tab: flickr kbp comparison}
\end{table}

\subsection{Performance on Flickr30K Entities}\label{sec: flickr exp}

\textbf{Comparison in accuracy.} We first evaluate pure visual consistency branch's performance for weakly supervised grounding task. 
In Table~\ref{tab: flickr30k res}, with a hard KBP gate, visual consistency achieves grounding accuracy as 28.53\%, which is very close to GroundeR model. 
Then we introduce soft KBP gate into visual consistency branch, which brings 2.03\% increase in accuracy. 
This indicates that visual consistency, even alone, is capable of providing good performance in weakly supervised scenario.
According to~\cite{rohrbach2016grounding}, GroundeR model is actually a basic case of language consistency branch without a KBP gate. 
We first introduce a hard KBP gate into language consistency branch, which brings 3.42\% increase in grounding performance.
We then replace the hard KBP gate with a soft KBP gate, which brings an additional 1.14\% increase in performance.
This further validates the effectiveness of external knowledge in weakly supervised grounding problem.
Finally, we combine visual and language consistency, which is the full KAC Net. 
By applying a hard KBP gate, KAC Net achieves 37.41\% in accuracy. 
We then replace the hard KBP gate with a soft KBP gate. 
The KAC Net reaches 38.71\% in accuracy, which is a 9.78\% increase over the performance of GroundeR~\cite{rohrbach2016grounding}.
From Table~\ref{tab: flickr30k res}, we also find soft KBP gate achieves consistently better performance over hard KBP gate.

\begin{figure*}[ht]
\includegraphics[width=6.5in]{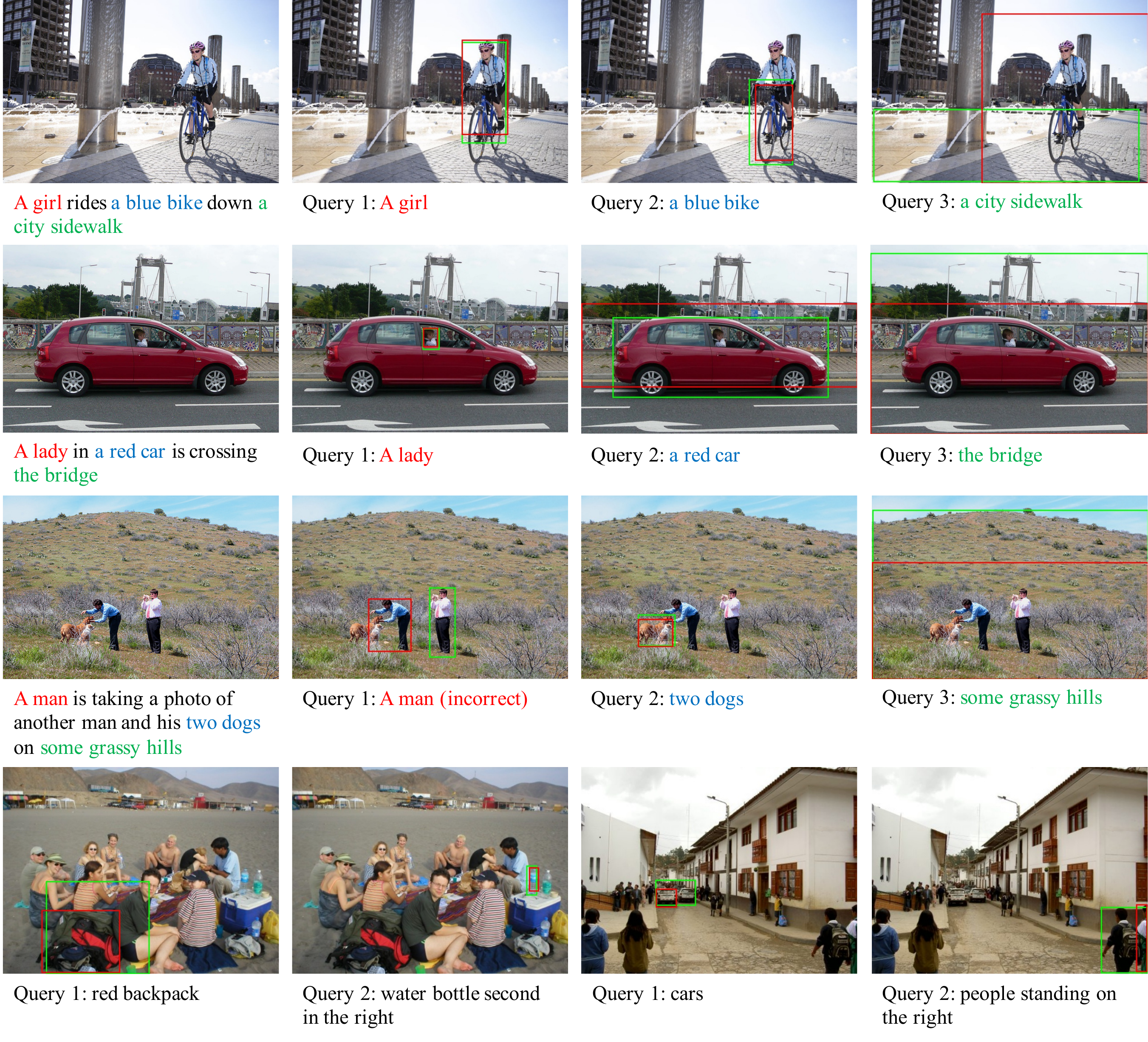}
\centering
\caption{Some phrase grounding results in Flickr30K Entities~\cite{plummer2015flickr30k} (first three rows) and Referit Game~\cite{KazemzadehOrdonezMattenBergEMNLP14} (forth row). We visualize ground truth bounding box and grounding result in green and red respectively. When query is not clear without further context information, KAC Net may ground reasonably incorrect objects (\emph{e.g.}, image in row three, column two).}\label{fig: demo flickr}
\end{figure*}

\begin{table*}[t]
  \centering
  \begin{tabular}{lcccccccc} \toprule
  Phrase Type & people & clothing & body parts & animals & vehicles & instruments & scene & other  \\ \midrule
  GroundeR (VGG\textsubscript{det})~\cite{rohrbach2016grounding} & 44.32 & \textbf{9.02} & 0.96 & 46.91 & 46.00 & 19.14 & 28.23 & 16.98 \\ \midrule
  LC + Soft KBP  & 55.23 & 4.21 & 2.49 & 67.18 & 54.50 & 11.73 & 37.37& 13.25 \\
  VC + Soft KBP & 51.56 & 5.33 & 2.87 & 58.11 & 51.50 & 20.01 & 26.86 & 12.63 \\
  KAC Net (Hard KBP) & 55.14 & 7.29 & 2.68 & 73.94 & 66.75 & 20.37 & 43.14 & 17.05 \\
  KAC Net (Soft KBP) & \textbf{58.42} & 7.63 & \textbf{2.97} & \textbf{77.80} & \textbf{69.00} & \textbf{20.37} & \textbf{43.53} & \textbf{17.05} \\ \bottomrule 
  \end{tabular}
  \vspace{1.5mm}
\caption{Phrase grounding performances for different phrase types defined in Flickr30K Entities. Accuracy is in percentage.}\label{tab: flickr30k detail res}
\end{table*}

\textbf{Detailed comparison.} Table~\ref{tab: flickr30k detail res} provides detailed weakly supervised grounding results based on the phrase type information for each query in Flickr30K Entities.
We can observe that KAC Net provides superior results in most categories. 
However, different models have different strength.
Language consistency with a soft KBP gate (LC+Soft KBP) is good at localizing ``people'', ``animal'' and ``vehicles'', with 10.91\%, 20.27\% and 8.5\% increase in accuracy compared to GroundeR model.
Compared to language consistency, visual consistency (VC+Soft KBP) is better at localizing ``clothing'', ``body parts'' and ``instruments'', with 1.12\%, 0.38\% and 8.28\% increase. 
However, for other categories, visual consistency branch achieves inferior performances.
By incorporating both visual and language consistency, KAC Net observes consistent improvement in all categories except for the category ``clothing''. 
With a soft KBP gate, KAC Net achieves 14.10\%, 23.00\% and 30.89\% increase in localizing ``people'', ``vehicles'' and  ``animals''. 
However, KAC Net also has 1.39\% drop in accuracy of localizing ``clothing''. 
This may be because ``clothing'' is usually on ``people''. 
In this case, there is high chance for a grounding system to classify ``clothing'' into ``people'' by mistake. 
Besides, ``clothing'' does not have corresponding categories in the external knowledge. 

\textbf{Knowledge representation.} To validate the effectiveness of external knowledge, we also evaluate KAC Net's performance using distributions predicted by VGG Network pre-trained on PASCAL VOC 2007~\cite{pascal-voc-2007} image classification. 
In Table~\ref{tab: flickr kbp comparison}, we observe that applying external knowledge achieves consistent improvement in grounding performance compared to GroundeR~\cite{rohrbach2016grounding} model. 
However, knowledge from MSCOCO~\cite{lin2014microsoft} image classification achieves a slight increase in accuracy compared to that from PASCAL VOC 2007~\cite{pascal-voc-2007} image classification. 
This may be because MSCOCO contains more categories of objects, and so may be more accurate in describing the proposal's relativeness to the query.




\begin{table}[t]
  \centering
  \begin{tabular}{lc} \toprule
  Approach & Accuracy (\%) \\ \midrule
  \textbf{Compared approaches} & \\
  LRCN~\cite{donahue2015long} & 8.59 \\
  Caffe-7K~\cite{guadarrama2014open} & 10.38 \\
  GroundeR~\cite{rohrbach2016grounding} (LC) (VGG\textsubscript{cls}) & 10.70 \\ \midrule
  \textbf{Our approaches} & \\
  LC + Hard KBP (VGG\textsubscript{cls}) & 13.02 \\ 
  LC + Soft KBP (VGG\textsubscript{cls}) & 13.97 \\
  KAC Net + Hard KBP (VGG\textsubscript{cls}) & 14.68 \\
  KAC Net + Soft KBP (VGG\textsubscript{cls}) & \textbf{15.83} \\
  \bottomrule
  \end{tabular}
  \vspace{1.5mm}
\caption{Different models' performance on Referit Game. We leverage knowledge from MSCOCO~\cite{lin2014microsoft} classification task.}\label{tab: referit res}
\end{table}

\begin{table}[t]
  \centering
  \begin{tabular}{|l|c|c|} \hline
  Knowledge & PASCAL VOC~\cite{pascal-voc-2007} & MSCOCO~\cite{lin2014microsoft} \\ \hline
  Hard KBP & 12.04 & 14.68 \\ \hline
  Soft KBP & 13.38 & 15.83
  \\\hline 
  \end{tabular}
  \vspace{1.5mm}
\caption{Comparison of KAC Net using different KBP gates and external knowledge on ReferitGame. Accuracy is in \%.}\label{tab: referit kbp comparison}
\end{table}

\subsection{Performance on Referit Game}\label{sec: referit exp}

\textbf{Comparison in accuracy.} Following~\cite{rohrbach2016grounding}, we adopt EdgeBoxes~\cite{zitnick2014edge} as a proposal generator. 
As shown in Table~\ref{tab: referit res}, by introducing KBP gate, KAC Net achieves 2.32\% (Hard KBP) and 3.27\% (Soft KBP) increase compared to state-of-the-art GroundeR~\cite{rohrbach2016grounding} model.
We observe using soft KBP gate achieves a slight increase in performance than hard KBP gate.
When KAC Net incorporates both visual and language consistency, it achieves another 1.66\% and 1.86\% increase compared to language consistency branch with hard and soft KBP respectively.
The full model achieves 15.83\% grounding accuracy, with 5.13\% increase over the GroundeR model.

\textbf{Knowledge representation.} Similar to Flickr30K Entities, we also evaluate KAC Net's performance using knowledge from PASCAL VOC 2007~\cite{pascal-voc-2007} image classification task. 
In Table~\ref{tab: referit kbp comparison}, we observe applying external learned from MSCOCO~\cite{lin2014microsoft} image classification achieves better performance than that from PASCAL VOC 2007~\cite{pascal-voc-2007}.
However, both knowledge representations help achieve increase in grounding accuracy over the state-of-the-art model.

\subsection{Discussion} 
To further explore KAC Net performance on different types of queries, we define queries with / without words in MSCOCO categories as ``Type A'' and ``Type B'' respectively. In Tables~\ref{tab: fairness flickr},~\ref{tab: fairness referit}, we evaluate two more compared methods: soft KBP only and pre-trained GroundeR~\cite{rohrbach2016grounding} with soft KBP (denoted as ``G + KBP'') on both Flickr30K Entities~\cite{plummer2015flickr30k} and Referit Game~\cite{KazemzadehOrdonezMattenBergEMNLP14} datasets.

From Tables~\ref{tab: fairness flickr},~\ref{tab: fairness referit}, pre-trained GroundeR shows a performance boost by adopting KBP. However, after end-to-end training (LC+KBP) and applying visual consistency part, KAC Net still outperforms state-of-the-art methods by a significant margin. 
These results also show the generalizability of KAC Net, with more details in the supplemental file.

\subsection{Qualitative Results}

We visualize some of KAC Net's grounding results on Flickr30K Entities and Referit Game datasets for qualitative evaluation in Fig.~\ref{fig: demo flickr}.
For Flickr30K Entities, we first show the image description where the query phrases come from, then show the grounding results and ground truth objects in red and green bounding boxes respectively.
For Referit Game, each query is independent with no common image descriptions, we visualize two example images with two queries in the third row of Fig.~\ref{fig: demo flickr}.

We find KAC Net is strong in recognizing people (``a girl'' in the first row) and vehicle (``cars'' in the third row), and is able to ground complex queries (``water bottle second in the right'' in the third row), which is also validated in Table~\ref{tab: flickr30k detail res}. 
   However, since KAC Net takes only single query phrase as input, it is unable to make use of context, such as in the example of ``a man'' in the third row of Fig.~\ref{fig: demo flickr}.\vspace{-1.5mm}



\begin{table}[t]
  \centering
  \begin{tabular}{l*3c} 
  \toprule
  {} & Type A & Type B & All \\
  \# queries & 1762 & 15757 & 17519 \\
  \midrule
  Soft KBP & 37.26 & 19.77 & 21.53 \\
  GroundeR & 26.54 & 29.19 & 28.93 \\
  G + KBP & 41.03 & 32.17 & 33.06 \\
  LC + KBP & 42.13 & 33.44 & 34.31 \\
  KAC Net & 45.66 & 37.93 & 38.71 \\
  \bottomrule
  \end{tabular}
  \vspace{1.0mm}
\caption{Different methods on Flickr30K Entities~\cite{plummer2015flickr30k} for two types of queries. Accuracy is in \%.}\label{tab: fairness flickr}
\end{table}

\begin{table}[t]
  \centering
  \begin{tabular}{l*3c} 
  \toprule
  {} & Type A & Type B & All \\
  \# queries & 8275 & 51796 & 60071 \\
  \midrule
  Soft KBP & 12.88 & 7.74 & 8.45 \\
  GroundeR & 7.29 & 11.24 & 10.70 \\
  G + KBP & 14.16 & 12.56 & 12.78 \\
  LC + KBP & 15.28 & 13.76 & 13.97 \\
  KAC Net & 18.36 & 15.43 & 15.83 \\
  \bottomrule
  \end{tabular}
  \vspace{1.0mm}
\caption{Different methods on Referit Game~\cite{KazemzadehOrdonezMattenBergEMNLP14} for two types of queries. Accuracy is in \%.}\label{tab: fairness referit}\vspace{-2.5mm}
\end{table}

\section{Conclusion}

We proposed a novel Knowledge Aided Consistency Network (KAC Net) to address the weakly supervised grounding task. 
KAC Net applies both visual and language consistency to guide the training and leverages free complementary knowledge to boost performance.
Experiments show KAC Net provides a significant improvement in performance compared to state-of-the-arts, with 9.78\% and 5.13\% increase in accuracy on Flickr30K Entities~\cite{plummer2015flickr30k} and Referit Game~\cite{KazemzadehOrdonezMattenBergEMNLP14} datasets respectively.

\section*{Acknowledgements}
This paper is based, in part, on research sponsored by the Air Force Research Laboratory and the Defense Advanced Research Projects Agency under agreement number FA8750-16-2-0204. 
The U.S. Government is authorized to reproduce and distribute reprints for Governmental purposes notwithstanding any copyright notation thereon. The views and conclusions contained herein are those of the authors and should not be interpreted as necessarily representing the official policies or endorsements, either expressed or implied, of the Air Force Research Laboratory and the Defense Advanced Research Projects Agency or the U.S. Government.

{\small
\bibliographystyle{ieee}
\bibliography{egbib}
}

\end{document}